\documentclass[submit,ams]{jsme-tj}
\AtBeginDvi{}
\usepackage{amssymb}
\usepackage[authoryear,round,sort&compress]{natbib}

\newcommand{\gv}[1]{\ensuremath{\boldsymbol{#1}}}
\newcommand{\dgv}[1]{\dot{\gv #1}}

\newcommand{\ep}{\gv{\omega}}
\newcommand{\ii}{{\gv{i}}}
\newcommand{\ee}{{\gv{e}}}
\newcommand{\oo}{{\gv{o}}}

\newcommand{\II}{{\Bbb I}}
\newcommand{\JJ}{{\Bbb J}}

\newcommand{\sgn}{{\mathop{\text{sgn}}}}
\newcommand{\trap}{{\mathop{\text{trap}}}}

\newcommand{\upd}{u^{\text{pd}}}
\newcommand{\uic}{u^{\text{ic}}}
\newcommand{\udic}{u^{\text{dic}}}

\newcommand{\Upd}{\gv u^{\text{pd}}}
\newcommand{\Uic}{\gv u^{\text{ic}}}

\newcommand{\uimp}{I_{\Delta\tau}}
\newcommand{\leqn}[1]{\label{eqn:#1}}
\newcommand{\reqn}[1]{(\ref{eqn:#1})}
\newcommand{\lfig}[1]{\label{fig:#1}}
\newcommand{\rfig}[1]{Fig.\ \ref{fig:#1}}
\newcommand{\Rfig}[1]{Figure\ \ref{fig:#1}}
\newcommand{\ltab}[1]{\label{tab:#1}}
\newcommand{\rtab}[1]{Table\ \ref{tab:#1}}
\newcommand{\lsec}[1]{\label{sec:#1}}
\newcommand{\rsec}[1]{Section\ \ref{sec:#1}}

\title{Artificial Wrestling: A Dynamical Formulation of Autonomous
Agents Fighting in a Coupled Inverted Pendula Framework}

\vol{1}
\no{1}
\field{XXX}
\received{2014/3/3}{XX-XXXX}

\author{A}{Katsutoshi YOSHIDA}
\author{A}{Shigeki MATSUMOTO}
\author{B}{Yoichi MATSUE}

\affiliate{A}%
{Department of Mechanical and Intelligent Engineering, Utsunomiya University}%
{7-1-2 Yoto, Utsunomiya, Tochigi, 321-8585, Japan}%
[yoshidak@cc.utsunomiya-u.ac.jp]%
\affiliate{B}{Yuhara Mfg Co., Ltd}%
{1256 Ujiie, Sakura, Tochigi 329-1311, Japan}

\begin{document}

\maketitle

\begin{abstract}
We develop autonomous agents fighting with each other, inspired by human
wrestling. For this purpose, we propose a coupled inverted pendula (CIP)
framework in which: 1) tips of two inverted pendulums are linked by a
connection rod, 2) each pendulum is primarily stabilized by a
PD-controller, 3) and is additionally equipped with an intelligent
controller. Based on this framework, we dynamically formulate an
intelligent controller designed to store dynamical correspondence from
initial states to final states of the CIP model, to
receive state vectors of the model, and to output impulsive control
forces to produce desired final states of the model. Developing a
quantized and reduced order design of this controller, we have a
practical control procedure based on an off-line learning
method. We then conduct numerical simulations to investigate
individual performance of the intelligent controller, showing that the
performance can be improved by adding a delay element into the
intelligent controller. The result shows that the performance depends
not only on quantization resolutions of learning data but also on delay
time of the delay element.  Finally, we install the intelligent
controllers into both pendulums in the proposed framework to demonstrate
autonomous competitive behavior between inverted pendulums.
\end{abstract}


\section{Introduction}
Wrestling seems to be composed artificially of two mechanical agents
maintaining their balance, coupled via mechanical interactions such as
contact, connection, collision, etc., and equipped with intelligent
controllers competitive with each other.  In this paper, we 
develop a simple model to create such competitive agents.  For
this purpose, we propose a coupled inverted pendula (CIP) framework
in which: 1) tips of two inverted pendulums are linked
by a connection rod, 2) each pendulum is primarily stabilized by a
PD-controller, 3) and is additionally equipped with an intelligent
controller that individually generates a series of impulsive internal
forces to achieve its own desired final states based on knowledge of
correspondence from initial states to the final states.

In general, multiple agents can exhibit competitive and cooperative dynamics
when sharing common resources and environments. Historically, early
mathematical insights into such mutual interactions seem to have
appeared in the filed of mathematical ecology \citep{Hofbauer1998} in
which population dynamics of different species sharing a common
environment is described by a system of coupled nonlinear differential
equations such as the Lotka-Volterra equation.  
Contrary to the ecosystem in which the medium of interaction is given by
environments, in our CIP framework, the medium of interaction is given
by a mechanical structure.  In our previous study \citep{cip2008}, we
have already demonstrated that the CIP model, even without intelligent
controllers, can produce competitive dynamics comparable to that in the
ecosystem such as coexistence and dominance by assigning competitive
meanings to the stable equilibriums of the CIP model.  Although quite
similar mechanical models have been considered in the field of multiple
manipulator systems \citep{Nakamura1987,Ping1993,Panwar2012216}, they
have only focused on cooperative dynamics because of their aim at
developing coordinated motions in those systems.

In our study mentioned above \citep{cip2008}, each pendulum is
PD-controlled to be bistable at the top and bottom dead points such that
the coupled system produces quadra-stability. In absence of additional
inputs, this system converges into one of the four stable positions
(equilibriums) depending upon the initial conditions.  We then applied a
single impulsive force to one of the pendulums to generate switching
behavior from a given stable position to a desired stable position and
considered it as a prototype of fighting-like behavior.  In this
prototype, however, the behavior is exactly determined by the initial
position and strength of the impulse because of uniqueness of solution of
differential equation. Therefore, it is quite hard to say that this
first prototype is comparable to the wrestling players who seek how to
generate internal forces to achieve desired final positions in autonomous
ways.

In order to build such autonomous agents fighting with each other,
a certain intelligent motion controller is required.
On such controllers, extensive research has been
conducted in the field of multi-robot systems \citep{Pagello1999,Maravall2013}.
The major issue in this field appears to be how to obtain cooperative
group dynamics of robots both in algorithm-based approaches \citep{Stone2000}
and in differential-equation-based approaches
\citep{Nakamura1987,Ping1993,Panwar2012216}.

On the other hand, competitive group behavior seems to have been studied
mainly based on algorithm-based approaches.
For example, \citet{Nelson2004135} studied an evolutionary
controller to investigate a form of reinforcement learning that makes
use of competitive tournaments of games (robot capture the flag) played
by individuals in a population of neural controllers.
Moreover, \citet{Wu201366} developed rule or knowledge-based
techniques to analyze strategy in robot soccer game.
In this way, in contrast to our approach \citep{cip2008},
differential-equation-based techniques are not always essential in these
studies because they seek step-by-step algorithms predicting free space
determined by other robots.

In this paper, we first introduce the CIP framework to describe the
competitive behavior in differential-equation- based manners. Next, we
develop a competitive intelligent controller that receives state
vectors of the CIP model and outputs impulsive control forces to produce
desired final states. After evaluating individual performance of the
intelligent controller and investigating how to improve the performance,
we will demonstrate autonomous competitive behavior between two inverted
pendulums equipped with the proposed intelligent controllers.

\section{Coupled inverted pendula framework}

\subsection{Coupled inverted pendula}

In order to create wrestler-like mechanical agents maintaining their
balance while being coupled mechanically with each other, we consider a
CIP model \citep{cip2008} as shown in \rfig{CIP}.  Each inverted
pendulum consists of a cart moving along the horizontal floor ($Y=0$)
and a simple pendulum rotating about a point on the cart.  For
simplicity, a common physical specification is given to both of the
pendulums where $m_\theta$ is a mass and $r$ is a length of the
pendulum, and $m_x$ is a mass of the cart. Linking the tips of the
pendulums with a viscoelastic connection rod of length $w$, we obtain
the CIP model in \rfig{CIP} where $T_i$ is an input torque on $\theta_i$,
$\gv f_i$ is a reaction force acting on the tip of $i$th pendulum, and $k_w$
and $c_w$ are a spring coefficient and a viscous friction coefficient
of the rod respectively. We assume that a mass of the rod is negligible.  As
configuration of this linkage is uniquely determined by the four
variables: horizontal displacements of the carts $x_1,x_2$ and slant
angles of the pendulums $\theta_1,\theta_2$, the dynamics of this
linkage is described by the eight-dimensional state vector:
\begin{equation}
 \gv x := (\gv x_1^T,\gv x_2^T)^T
  ,\quad
 \gv x_i := \big(x_i,\dot x_i,\theta_i,\dot \theta_i\big)^T
 \quad(i=1,2),
 \leqn{statevector}
\end{equation}
where $A^T$ denotes the transpose of a matrix $A$.
According to Lagrangian mechanics and assuming viscous friction forces
$c_x\dot x_i$ and $c_\theta\dot\theta_i$ on $x_i$ and $\theta_i$ respectively, we
obtain equations of motion (EOM) of the CIP model in \rfig{CIP} as
follows:
\begin{equation}
 \begin{cases}
  (m_x+m_\theta)\ddot{x}_{i} + ( m_\theta r\cos\theta_{i} )\ddot{\theta}_{i} 
  - m_\theta r\dot{\theta}_{i}^{2}\sin\theta_{i}
  = -c_{x}\dot{x}_{i} + (1,0)\gv f_i, 
  \\
  (m_\theta r\cos\theta_{i})\ddot{x}_{i} + (m_\theta r^{2})\ddot{\theta}_{i} 
	- m_\theta gr\sin\theta_{i} 	
  = - c_{\theta}\dot{\theta}_{i} 
    +r(\cos\theta_i,-\sin\theta_i)\gv f_i + T_i 
  \quad (i=1,2),
 \end{cases}
 \leqn{EOM}
\end{equation}
where $\dot X :=dX/dt$.

\begin{figure}[t]
\centering
\begin{minipage}[t]{.55\hsize}
\centering
\includegraphics[width=\hsize]{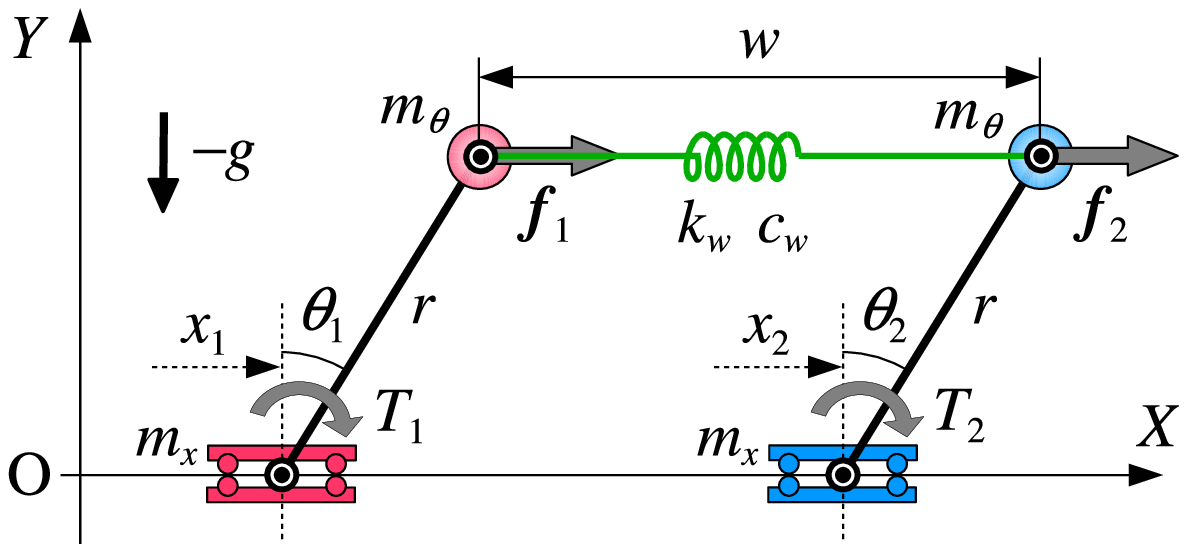}
\caption{Coupled inverted pendula model via viscoelastic connection.}
\lfig{CIP}
\end{minipage}\hfill
\begin{minipage}[t]{.4\hsize}
\centering
\includegraphics[width=\hsize]{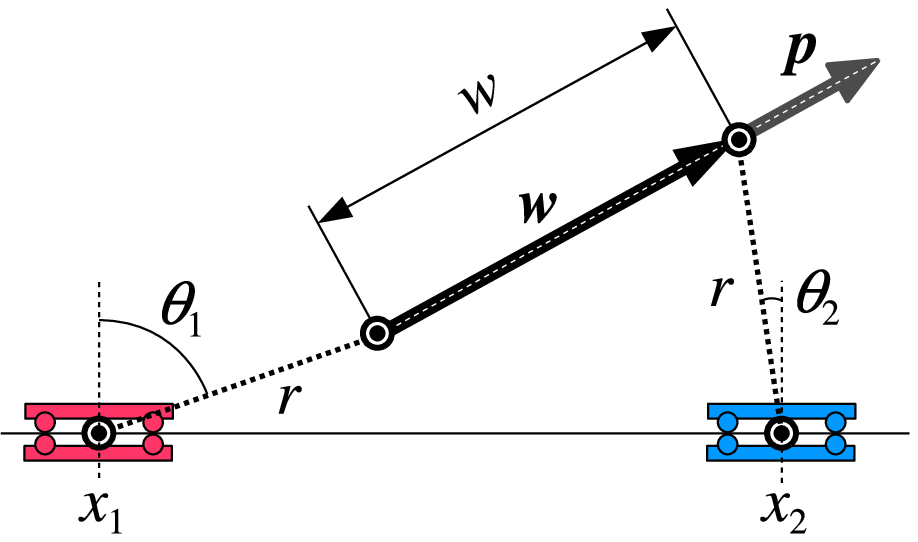}
\caption{Reaction force $\gv p$ from the connection rod.}
\lfig{CIP-rod}
\end{minipage}
\end{figure}

\subsection{Reaction force from the connection rod}

We calculate the reaction force, say $\gv p$, from the connection rod 
as shown in \rfig{CIP-rod}. The displacement vector $\gv w$ from the
left-hand tip to the right-hand tip of pendulums is given by
%
%
\begin{equation}
 \gv w = 
  \begin{bmatrix}
   w_X \\ w_Y
  \end{bmatrix}
  := \gv X_2 -\gv X_1
  ,\quad
  \gv X_i = 
  \begin{bmatrix}
   X_i \\ Y_i
  \end{bmatrix}
  =
  \begin{bmatrix}
   x_i + r\sin\theta_i\\
   r\cos\theta_i
  \end{bmatrix}
  \quad(i=1,2),
  \leqn{ww}
\end{equation}
and the length of rod is expressed as
\begin{equation}
 w = \|\gv w\| = \sqrt{
  \{(x_2-x_1) + r(\sin\theta_2-\sin\theta_1)\}^2
  +\{r(\cos\theta_2-\cos\theta_1)\}^2}
\end{equation}
with the time derivative
\begin{math}
 \dot w = (\dot w_X w_X + \dot w_Y w_Y)/w.
\end{math}
Then, we model viscoelasticity of the connection rod as
\begin{equation}
 p = \|\gv p\| := -k_w(w-w_0)-c_w\dot w,
 \leqn{p}
\end{equation}
where $w_0$ is a natural length of the connection rod. As the force
vector $\gv p$ is parallel to the displacement vector $\gv w$, we
have the reaction force:
\begin{equation}
 \gv p = (p/w)\gv w.
  \leqn{pp}
\end{equation}
Substituting $\gv p$ into the EOM \reqn{EOM} through,
\begin{equation}
 \gv f_1 = -\gv p
  ,\quad 
 \gv f_2 = \gv p
 \quad\big(\,\text{or \;$\gv f_i=(-1)^i\gv p$}\;\big),
 \leqn{rod}
\end{equation}
we obtain an analytic expression of the CIP model shown in \rfig{CIP}
via the viscoelastic connection.

\subsection{Modeling floor}

In the previous study \citep{cip2008}, the pendulum can fall down freely
to the bottom dead point, in other word, there was no floor in the
previous CIP model. In that case, both of forward and backward falling
motions converge to the same equilibrium of the model so that orbital
information is required in order to detect the direction of falling.
Not only to simplify the detection process but also to develop more
realistic simulator of wrestling, we introduce the floor model to the
CIP model in the following manner.

Based on penalty methods \citep{Moore1988}, we first model a normal
force $R_i$ from the floor ($Y=0$) acting on the tip of $i$th
pendulum as
\begin{equation}
 R_i = U(-Y_i)\{-k_fY_i-c_f\dot Y_i\},
 \leqn{reaction}
\end{equation}
where $Y_i$ is a height of the $i$th tip from the floor in \reqn{ww},
$U(\,\cdot\,)$ is a unit step function, and $k_f,c_f$ are viscoelastic
parameters representing property of the reaction.  In practice, in order
to avoid numerical errors, we approximate the step function with a
sigmoid function differentiable, defined by
\begin{equation}
 U_\sigma(s) := \left\{1+\exp(-\sigma s)\right\}^{-1},
  \leqn{sigmoid}
\end{equation}
where $\lim_{\sigma\to\infty}U_\sigma(s)=U(s)$ holds.

Furthermore, a Coulomb friction force
$F_i$ acting on the $i$th tip from the floor can be expressed as
\begin{equation}
 F_i = -\mu R_i\,\sgn(\dot X_i),
\end{equation}
where $\mu$ is a friction coefficient, $\dot X_i$ is a relative horizontal
velocity of the $i$th tip from the floor, and $\sgn(\,\cdot\,)$ is a
unit signum function whose smooth approximation can be given by
\begin{math}
 \sgn(s) \approx \sgn_\sigma(s) := 2U_\sigma(s)-1.
\end{math}

Therefore, the CIP model via the viscoelastic connection in \reqn{rod}
on the floor can be obtained by substituting
\begin{equation}
 \gv f_i = (-1)^i \gv p + 
  (F_i, R_i)^T
  \quad (i=1,2)
  \leqn{qq}
\end{equation}
into the EOM in \reqn{EOM}.

\subsection{Standing control with falling}

The CIP framework in absence of intelligent controllers is completed by
giving dynamical meanings of winning and losing to states of the CIP
model.  To this end, we begin with developing a feedback controller by
which each inverted pendulum on the floor forms three stable
equilibriums: $\theta_i=0$ for standing or winning and
$\theta_i=\pm\pi/2$ for falling or losing.  This can be done by introducing
a feedback controller in the following form:
\begin{equation}
 T_i = \upd_i
  :=\trap_\alpha(\theta_i;\Delta\theta)
    \{-K_p\theta_i-K_d\dot\theta_i\}\quad (i=1,2),
  \leqn{upd} 
\end{equation}
where 
\begin{equation}
 \trap_\alpha(\theta;\Delta\theta) := 
  U_\alpha(\theta+\Delta\theta)\cdot U_\alpha(-\theta+\Delta\theta)
\end{equation}
is a smooth trapezoidal function of unit height and centered at
$\theta=0$ made of a product of the sigmoid function in \reqn{sigmoid},
$\Delta\theta>0$ is a half width of the trapezoidal shape, and the shape
is getting steeper as $\alpha$ increases.

It follows from the deadband characteristics in \reqn{upd} that $\upd_i$
simply acts as a PD controller within the limit $|\theta_i|<\Delta\theta$
while it rapidly cuts off the output outside of the interval.
Therefore, appropriate setting of the gains $K_p,K_d$ make it possible
for the $i$th pendulum to be stabilized about the standing position
$\theta_i=0$ while to be falling down to the floor when $|\theta_i|$
exceeds the given limit $\Delta\theta$.

It is worthy to note that in the field of gerontology and related
fields, human standing (or falling) limits comparable to the threshold
$\Delta\theta$ have been measured by the {\em functional reach test}
\citep{Duncan1990} in which the difference between arm length and maximal
forward reach of human subjects is measured to evaluate risk of falls of them.

\subsection{CIP framework}
\lsec{CIPframework}

In what follows, we compare the inverted pendulums in this model to
wrestler-like agents maintaining their standing balance.
Since each agent with the standing control in \reqn{upd} has the three
stable equilibriums, a pair of the agents being coupled with each other
under the suitable conditions can produce $3\times3=9$ stable equilibriums:
\begin{equation}
 \ep_i := \lim_{t\to\infty} \gv x(t)
 =
 \big(
 \bar x_1,0,
 \bar\theta_1,0,
 \bar x_2,0,
 \bar\theta_2,0
 \big)^T
  \quad(i=1,\cdots,9),
\end{equation}
as shown in \rfig{ep} when equating horizontal translations of final
position $\bar x_1,\bar x_2$ without loss of generality. Namely, the
components $x_1(t),x_2(t)$ of the solution $\gv x(t)$ of \reqn{EOM}
are not stable asymptotically but neutrally because no restoring forces
on $x_1(t),x_2(t)$ are assumed by definition.  It also should be noted
that due to the penalty method in \reqn{reaction}, the gravity
force makes the equilibrium $\bar\theta_i$ on the floor slightly exceed
the floor, i.e., $|\bar\theta_i|-\pi/2>0$, but we formally
denote $\theta_i=\bar\theta_i=\pm\pi/2$ because this slight exceedance
only affects almost converged states and does not change the
correspondence from the initial to final states.

\begin{figure}[t]
\centering
\begin{minipage}[t]{.47\hsize}
\centering
\includegraphics[width=\hsize]{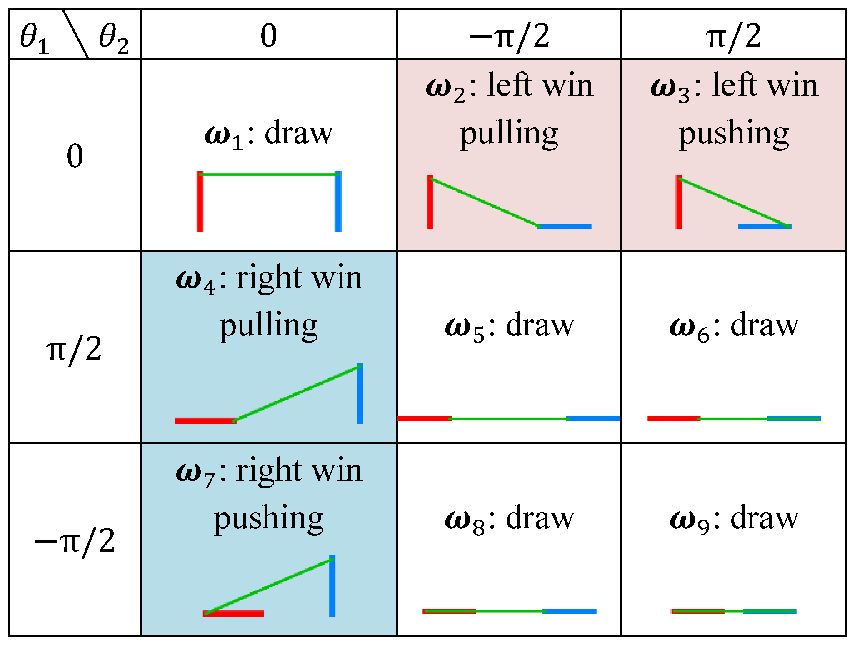}
\caption{Competitive interpretation of the equilibriums.}
\lfig{ep}
\end{minipage}\hfill
\begin{minipage}[t]{.44\hsize}
\centering
\includegraphics[width=\hsize]{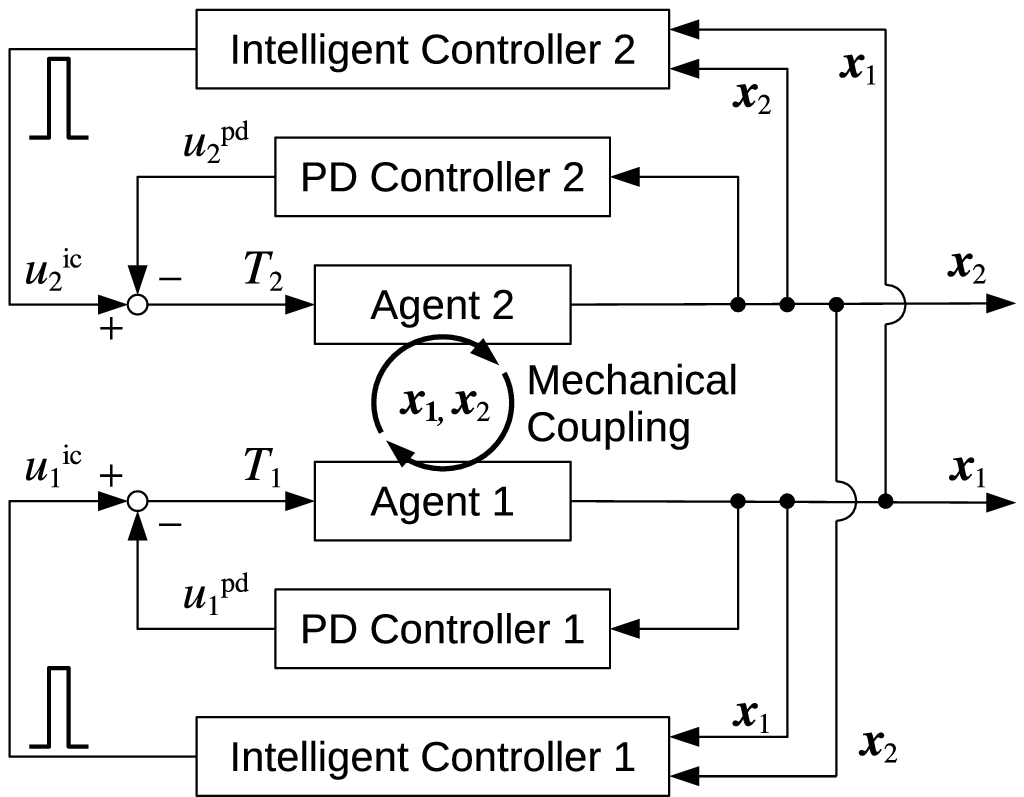}
\caption{Architecture of the proposed control system
to generate\\ competitive motions.}
\lfig{CIPwithControl}
\end{minipage}
\end{figure}

We then attach competitive meanings to the nine equilibriums as listed in
\rfig{ep} in which the agent that remains standing is regarded as a
winner. Eventually, we have the {\it CIP framework} composed of the set of
(A) and (B):
\begin{itemize}
 \item[(A)] The CIP model: the system of equations defined in \reqn{EOM},
	    \reqn{qq} and \reqn{upd}.
 \item[(B)] The win-loss matrix: the competitive interpretation of the
	    nine equilibriums defined in \rfig{ep}.
\end{itemize}

\section{Intelligent controller}

Based on the CIP framework in \rsec{CIPframework}, we develop an
intelligent controller (IC) to produce desired final positions in
\rfig{ep} from given initial states of CIP by generating certain
impulsive forces.

\subsection{Problem setting and requirements}

According to the definition of the state vector $\gv x=(\gv x_1^T,\gv x_2^T)^T$
in \reqn{statevector}, the CIP model in \reqn{EOM} can be expressed as
an eight-dimensional dynamical system:
\begin{equation}
  \dgv x = \gv f(\gv x,\gv T)
   ,\quad \gv x(0)=\gv x_0
   ,\quad \gv T:=
   (T_1,T_2)^T
   \leqn{CIP8D}
\end{equation}
that can be divided into a pair of four-dimensional subsystems:
\begin{equation}
 \begin{cases}
  \dgv x_1 = \gv f_1(\gv x_1,\gv x_2,T_1),\\
  \dgv x_2 = \gv f_2(\gv x_2,\gv x_1,T_2).
 \end{cases}
   \leqn{N-DS2}
\end{equation}

We introduce the IC by adding an intelligent control
input $\uic_i$ to the torque $T_i$ of CIP model as
\begin{equation}
 \gv T:=\Upd+\Uic
  =\big(\upd_1,\upd_2\big)^T + \big(\uic_1,\uic_2\big)^T,
\end{equation}
where $\upd_i$ is the standing control input already given in \reqn{upd}.

In the present study, the input-output relationship around 
$\uic_i$ is designed as shown in \rfig{CIPwithControl}, in which 
$\uic_i$ receives all the state vectors $\gv x_1,\gv x_2$ and
outputs a series of impulsive forces given by
\begin{equation}
 \uic_i(t) := \sum_{j=1}^{N}P_i\,\uimp(t-t_i^j),
  \leqn{uic}
\end{equation}
where
\begin{equation}
 \uimp(t) =
  \begin{cases}
   (\Delta\tau)^{-1} & (0\leq t < \Delta\tau),\\
   0 & (\text{otherwise})
  \end{cases}
  \leqn{imp}
\end{equation}
is a rectangular function of unit area of width $\Delta\tau\ll 1$, $P_i$
is an angular impulse of the input torque $\uic_i(t)$, and
$\{t_i^1,\cdots,t_i^N\}$ is a series of rise time satisfying
\begin{equation}
 t_i^1 < t_i^2 < \cdots < t_i^N,\quad 
  \max_{j,k}|t_i^j-t_i^k|\geq\tau_G\geq\Delta\tau,
  \leqn{tau_G}
\end{equation}
where $\tau_G$ is a relaxation time to avoid overlapped outputs.

In practical implementation, the rise times $t_i^1,\cdots,t_i^N$ are
supposed to be determined sequentially by a real time architecture
described in \rfig{IC}, which is composed of three components, a
classifier $C$, a selector $S_\JJ$, and an impulse generator $G$.

\subsection{Classifier $C$}
\lsec{classifier}

We define the classifier $C$ as a function from a state vector $\gv
x=(\gv x_1^T,\gv x_2^T)^T$ at the time $t=t_0$, say $\gv x(t_0)=\gv
\xi_0$, to an index number $\nu$ of equilibrium $\ep_{\nu}$. The
function $C$ takes the value $C(\gv\xi_0)=\nu$ if a solution of the
following system:
\begin{equation}
  \dgv x = \gv f(\gv x,\gv T)
   ,\quad \gv x(t_0)=\gv \xi_0
   ,\quad \gv T:= \Upd + 
   \begin{bmatrix}
    P_1\,\uimp(t-t_0) \\ 0
   \end{bmatrix}
   \leqn{imp-res}
\end{equation}
converges to the equilibrium $\ep_\nu$.  
From this definition in which a single impulse at $t=t_0$ is applied
only on the left-hand agent,
it is implied that the classifier $C$ is valid only in absence of
additional inputs, in other words, it can fail to return correct
equilibriums for more general cases of input as in \reqn{uic} where both
agents can produce impulsive forces for their own decisions.  Despite
that, we will proceed with a discussion to build a first prototype of
artificial wrestling.

Consider the transition operator of a solution of \reqn{imp-res} as
\begin{equation}
 \gv x(t) := \gv\phi_t(\gv\xi_0,\gv T),
 \leqn{tranop}
\end{equation}
and define a set of the initial state $\gv\xi_0$ approaching the
equilibrium $\ep_i$ as
\begin{equation}
 \Phi_i :=\Big\{ 
  \gv\xi_0\in R^8
  \;\Big|\; 
  \lim_{t\to\infty} \gv\phi_t(\gv\xi_0,\gv T) = \gv\ep_i
  \Big\}.
  \leqn{basin}
\end{equation}
The set $\Phi_i$ is generally called a basin of attraction
\citep{ACS:ACS1195} or a reachable set \citep{Bayadi2013}.  It follows
from uniqueness of solution of initial value problem in \reqn{imp-res}
that the reachable set in \reqn{basin} satisfies,
\begin{equation}
 \Phi_i \cap \Phi_j = \emptyset\qquad(i\neq j).
\end{equation}
Thus, the classifier $C$ is obtained as a single-valued function:
\begin{equation}
 C\big(\gv\xi_0\big) :=
  \nu\quad\text{if}\quad \gv\xi_0\in \Phi_{\nu}.
  \leqn{classifier}
\end{equation}

In \rsec{approx}, we will discuss a numerical approximation of $C$
because explicit expressions of $C(\,\cdot\,)$ are hardly obtainable
from nonlinear systems such as \reqn{CIP8D} and also \reqn{imp-res}.

\subsection{Selector $S_\JJ$}

In the competitive problem in \rfig{ep}, some of the equilibriums are
selected depending upon strategies of the agent considered.  Such a
selection process can be modeled by a selector $S_\JJ$ given by
\begin{equation}
 \delta = S_\JJ(\nu) :=
  \begin{cases}
   1 & (\nu\in \JJ\subset \{1,\cdots,9\}),\\
   0 & (\text{otherwise}),
  \end{cases}
\end{equation}
where $\nu=C(\gv x)$ is an output of the classifier and $\JJ$ is a given
subset of indices of the equilibriums $\ep_1,\cdots,\ep_9$.  For
example, the trajectory $\gv x(t)$ in \reqn{imp-res} starting from an
initial state $\gv \xi_0$ satisfying
\begin{equation}
 (S_\JJ\circ C)(\gv\xi_0) := S_\JJ\Big(C(\gv\xi_0)\Big)=1
  \quad\text{for}\quad
  \JJ=\{2,3\}
\end{equation}
converges one of the two equilibriums $\ep_2$ and $\ep_3$.

\subsection{Impulse generator $G$}
\lsec{G}

The impulse generator $G$ is designed to receive the binary signal
$\delta(t)=S_\JJ(\nu(t))$ from the selector and output the unit impulse
$G(t)$ as shown in \rfig{IG}, which is composed of a two-input AND
gate and two timer functions $T_I$ and $T_G$. The timer $T_I$ produces a
unit impulse as
\begin{equation}
 G(t) = T_I(t):= \uimp(t-t_r),
\end{equation}
and the timer $T_G$ cuts off the binary signal $\delta(t)=S_\JJ(\nu(t))$
for a given relaxation time $\tau_G$ discussed in \reqn{tau_G} as
\begin{equation}
 T_G(t):=
  \begin{cases}
   0 & (t_r < t < t_r+\tau_G),\\
   1 & (\text{otherwise}),
  \end{cases}
\end{equation}
where $t_r$ is the rise time from 0 to 1 of the Boolean product
$\hat\delta(t)=S_\JJ(\nu(t)) \wedge T_G(t)$.  
Note that although the signal $\hat\delta(t)$ can be pulses of
infinitesimal width in this continuous time expression, this problem
will never arise in discrete time applications with digital computers.

\begin{figure}[t]
\centering
\begin{minipage}[t]{.45\hsize}
\centering
 \includegraphics[width=\hsize]{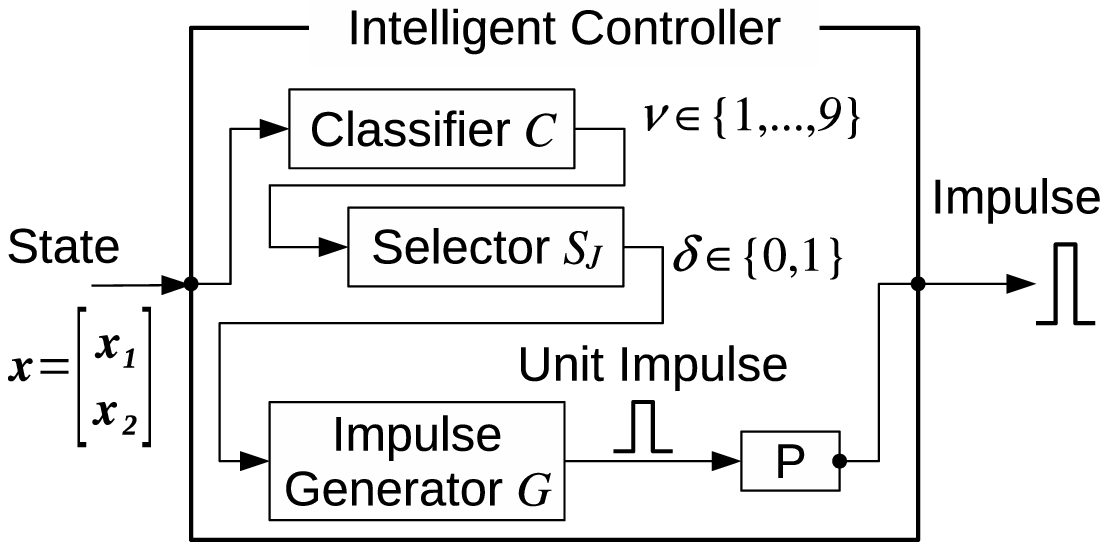}
 \caption{Intelligent controller.}
 \lfig{IC}
\end{minipage}\hfill
\begin{minipage}[t]{.53\hsize}
 \centering
 \includegraphics[width=\hsize]{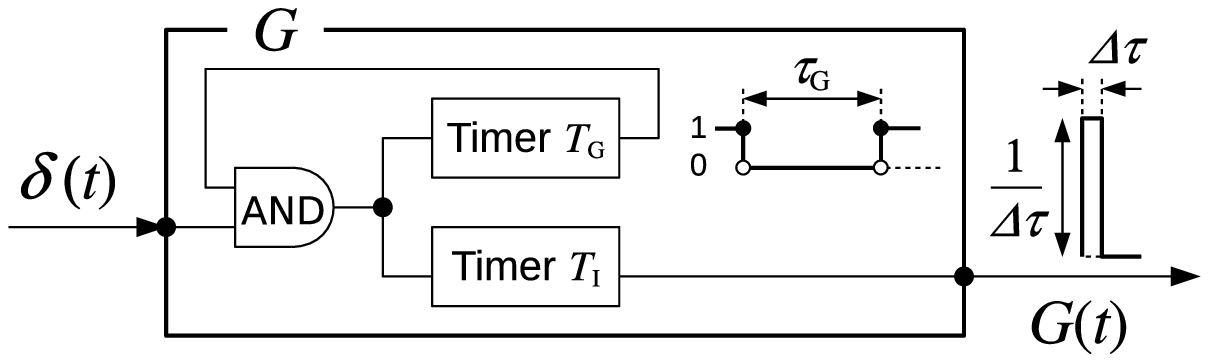}
 \caption{Impulse generator.}
 \lfig{IG}
\end{minipage}
\end{figure}

\subsection{Numerical approximation of the classifier $C$}
\lsec{approx}

Explicit expression of $C(\gv x)$ is hardly obtainable from nonlinear
systems such as \reqn{CIP8D}.  It seems that there are two types of
solutions: solving \reqn{CIP8D} in process or making a numerical table
of the mapping $C:\gv x\mapsto \nu$ in advance.  We take the latter
approach in the following manner.

For practical applications, we introduce a linear measurement equation:
\begin{equation}
 \gv y = H\gv x,\quad \gv y\in R^M,\gv x\in R^8,
 \leqn{measure-eq}
\end{equation}
where $M\leq 8$ and $H$ is a $M\times 8$ matrix of
rank $M$, and define a reduced-order reachable set in the following
sense:
\begin{equation}
 \Psi_i = H(\Phi_i) :=\Big\{ 
  \gv\eta_0\in R^M
  \;\big|\; 
  \gv\xi_0=h(\gv\eta_0)\in R^8
  ,\;
  \lim_{t\to\infty} \gv\phi_t(\gv\xi_0,\gv T) = \gv\ep_i
  \Big\},
  \leqn{basin-reduce}
\end{equation}
where $\gv x'=h(\gv y)$ ($\neq\gv x$ in general) is a certain
inverse of $\gv y=H\gv x$.  An actual example of $h$ will be given in
\rsec{rigid}.

Firstly, we take a $M$-dimensional cubic region $D$ of measuring range
within a direct sum of the reachable sets as
\begin{equation}
 \bigoplus_{i=1}^9 \Psi_i \supset D
 :=[a_1,b_1]\times\cdots\times[a_M,\cdots,b_M]
\end{equation}
where $\bigoplus$ represents a direct sum (disjoint union) and $[a,b]$
denotes an interval. We divide it into a direct sum of uniform subcubes
$D^\ii$ as
\begin{equation}
 D = \bigoplus\Big\{ 
  D^{\ii}
  \;\big|\; 
  \ii\in\II = [1,\cdots,m_1]\times\cdots\times[1,\cdots,m_M]
  \Big\},
\end{equation}
where $\II$ is a space of $M$-dimensional indices and $m_j$ is the number of
subcubes in the $j$th direction.  We then define center points of the
subcubes
\begin{math}
 \gv y^{\ii} \in D^{\ii}\; (\ii\in\II)
\end{math}
whose $j$th component is given by
\begin{equation}
 (\gv y^{\ii})_j := 
  a_j + \left((\ii)_j-\frac12\right)
  \frac{b_j-a_j}{m_j},
\end{equation}
where $(\gv v)_j$ denotes the $j$th component of a vector $\gv v$.

The setup above allows us to build a numerical method as follows:
\begin{subequations} 
\begin{enumerate}
 \item As an offline learning procedure, the mapping $\bar C$:
       \begin{equation}
	\quad\bar C(\ii) :=  \nu
	 \quad\text{if}\quad 
	 \lim_{t\to\infty} \gv \phi_t\big(h(\gv y^\ii),\gv T\big) = \ep_\nu,
	 \leqn{Capprox}
       \end{equation}
       is numerically stored by solving \reqn{imp-res} from
       $\gv\xi_0=h(\gv y^\ii)$.
 \item Within the IC in process, the classifier $C$ is quantized
       by $C^\ast$ as
       \begin{equation}
	\quad C(\gv x) \approx C^\ast(\gv x) :=
	 \bar C(\ii) 
	 \quad\text{for}\;\;
	 \ii
	 \;\;\text{such that}\;\;
	 H\gv x \in D^\ii.
       \end{equation}
\end{enumerate}
\leqn{algorithm}
\end{subequations}
In this method, accuracy of the classifier $C^\ast$ depends on the
dimension of measurement $M$, the size and placement of measuring range
$[a_j,b_j]$, and the resolution of quantization of reachable set $m_j$
($j=1,\cdots,M$).

In summary, we have obtained the IC for the
left-hand agent as a composed function of the quantized classifier
$C^\ast$, the selector $S_{\JJ}$, and the impulse generator $G$, 
which results in the closed-loop form:
\begin{equation}
 \uic_1 =\uic_1\big(\gv x(t);\JJ\big)
  := P_1\cdot(G\circ S_\JJ\circ C^\ast)\big(\gv x(t)\big).
\end{equation}

If $M=8$ and $\tau_G$ is sufficiently large, it is implied that the
solution $\gv x(t)$ in \reqn{imp-res} starting from any initial states
in $D$ undergoes an impulsive force at the time $t=t_0$ that $\uic_1$
decides autonomously and that it converges to the stable equilibriums
specified by $\JJ$, under the resolution limit, $\min_j(m_j) \to
\infty$.

\subsection{Reduced-order design of measurement for rigid connection}
\lsec{rigid}

The connection rod can behave like a rigid rod if the viscoelastic
parameters $k_w$, $c_w$ are sufficiently large.
For the rest of this paper, we restrict our problem to this nearly rigid
connection.
Although human wrestling involves more flexible interactions between
agents, this allows us substantially to reduce the computational efforts as follows.
In this case, dependency between displacements and velocities of the
left-hand and right-hand carts is imposed in the following form:
\begin{subequations} 
\begin{align}
 x_2 & = x_2(x_1,\theta_1,\theta_2)
 = x_1
  - r(\sin\theta_2-\sin\theta_1)
  + \sqrt{w_0^2 - r^2(\cos\theta_2-\cos\theta_1)^2},
 \\
 \dot x_2 & = \dot x_2(\dot x_1,\theta_1,\dot\theta_1,\theta_2,\dot\theta_2)
 = \dot x_1
  - r(\dot\theta_2\cos\theta_2-\dot\theta_1\cos\theta_1)
  +\frac{r^2(\cos\theta_2-\cos\theta_1)(\dot\theta_2\sin\theta_2-\dot\theta_1\sin\theta_1)}{\sqrt{w_0^2 - r^2(\cos\theta_2-\cos\theta_1)^2}},
\end{align}
\leqn{constraint}%
\end{subequations}%
where $w=w_0\gg 2r$ is a constant length of the rigid rod.
Then, a loss-less state feedback to IC can be done by the following
measurement:
\begin{subequations}
\begin{align}
 \gv y = H\gv x 
 = (x_1,\dot x_1,\theta_1,\dot\theta_1,\theta_2,\dot\theta_2)^T,
 \quad
 H = \left[\ee_1^{(6)},\ee_2^{(6)},\ee_3^{(6)},\ee_4^{(6)},
     \oo^{(6)},\oo^{(6)},\ee_5^{(6)},\ee_6^{(6)}\right],
\end{align}
where $\ee_i^{(d)}$, $\oo^{(d)}$ denote the $i$th standard basis vector
and the zero vector in Euclidean space $R^d$ respectively.  A
corresponding inverse satisfying the rigid constraint can be defined by
\begin{equation}
 h(\gv y):=H^{+}\gv y 
  + x_2(x_1,\theta_1,\theta_2)\ee_5^{(8)} 
  + \dot x_2(\dot x_1,\theta_1,\dot\theta_1,\theta_2,\dot\theta_2)\ee_6^{(8)},
\end{equation}
\end{subequations}
where $H^{+}$ is the Moore-Penrose pseudoinverse of $H$.  This
measurement reduces computational efforts to obtain the quantized
reachable sets of $\bar C$ in \reqn{Capprox} from $O(m^8)$ to $O(m^6)$ 
with respect to the resolution of quantization $m$.

In the following numerical examples, we perform a further reduction of
order given by
\begin{subequations}
\begin{align}
 &\gv y = H\gv x 
 = (\theta_1,\dot\theta_1,\theta_2,\dot\theta_2)^T
 ,\quad
 H = \left[\oo^{(4)},\oo^{(4)},\ee_1^{(4)},\ee_2^{(4)},
     \oo^{(4)},\oo^{(4)},\ee_3^{(4)},\ee_4^{(4)}\right],
 \\
 &h(\gv y):=H^{+}\gv y 
 + x_1\ee_1^{(8)} 
 + \dot x_1\ee_2^{(8)}
 + x_2(x_1,\theta_1,\theta_2)\ee_5^{(8)} 
 + \dot x_2(\dot x_1,\theta_1,\dot\theta_1,\theta_2,\dot\theta_2)\ee_6^{(8)}
 \quad\text{with $x_1=\dot x_1=0$}.
\end{align}
Although this measurement suffers complete loss of information about the
cart motion $x_1,\dot x_1$ (and $x_2,\dot x_2$ via \reqn{constraint}),
it reduces the computational efforts into $O(m^4)$. 
\leqn{measure4D}
\end{subequations}
In this paper, we employ this four-dimensional measurement in
priority to reducing the computational efforts. 
Moreover, in this four-dimensional measurement, the controller
$\uic_1(\gv x;\JJ)$ originally designed for the left-hand agent can
symmetrically be reused for the right-hand agent by a transformation:
\begin{equation}
 \uic_2(\gv x;\JJ') := \uic_1(\gv x';\JJ)
  ,\quad
  P_2 := -P_1
  ,\quad
  \gv x' := - (\gv x_2^T,\gv x_1^T)^T,
  \leqn{uic2}
\end{equation}
where $\JJ'=\{4,7\}$ for $\JJ=\{2,3\}$ due to the transpose of 
the $3\times3$ matrix of $\ep_i$ in \rfig{ep}.

\section{Numerical investigation}

We conduct numerical experiments to evaluate performance of the
four-dimensional IC of the measurement in \reqn{measure4D}. For
simplicity, in what follows, we set the resolution $m_j$ of quatization
to a common value $m$ for all $j$. The parameter values used in the
following examples are listed in \rtab{CIPpara}. The physical dimensions
$m_\theta,m_x$ and $r$ are roughly collected from the commercially
available inverted pendulum \citep{ZMP}. For numerical integration, the
fourth-order Runge-Kutta-Gill method is employed with the time step
$\Delta t$ listed in \rtab{ICpara}.

\begin{table}[t]
 \begin{minipage}[t]{.49\hsize}
 \centering
 \caption{Parameter setting of the CIP model.}
 \begin{tabular}{c@{\quad}ll}\hline
  \multicolumn{2}{l}{Parameters} & \multicolumn{1}{l}{Values}
  \\\hline
  $m_\theta$ & mass of pendulum & 0.68 kg
	  \\
  $m_x$ & mass of cart & 0.067 kg
	  \\
  $g$ & acceleration of gravity & 9.8 m/s$^2$
	  \\
  $r$ & length of pendulum & 0.3 m
	  \\
  $c_x$ & viscous coefficient along $x$ & $0.01$ Ns/m
	  \\
  $c_\theta$ & viscous coefficient about $\theta$ & $0.01$ Ns/m
	  \\
  $w_0$ & natural length of connection rod & 1 m
	  \\
  $k_w$ & spring coefficient of connection rod & $5000$ N/m
	  \\
  $c_w$ & viscous coefficient of connection rod & $50$ Ns/m
	  \\
  $k_f$ & spring coefficient of floor  & $500$ N/m
	  \\
  $c_f$ & viscous coefficient of floor & $10$ Ns/m
	  \\
  $\mu$ & Coulomb friction coefficient of floor & $0$ 
	  \\
  $\sigma$ & steepness of step function& $10^6$
	  \\
  $\alpha$ & steepness of trapezoidal function& $25$
	  \\
  $K_p$ & proportional gain of standing control & $1$
	  \\
  $K_d$ & derivative gain of standing control & $0.01$
	  \\
  $\Delta\theta$ & threshold of standing control & $\pi/6$ rad
	  \\
  \hline
 \end{tabular}
 \ltab{CIPpara}
 \end{minipage}
 \hfill
 \begin{minipage}[t]{.49\hsize}
 \centering
 \caption{Parameter setting of the four-dimensional IC\\and numerical integration.}
 \begin{tabular}{c@{\quad}ll}\hline
  \multicolumn{2}{l}{Parameters} & \multicolumn{1}{l}{Values}
  \\\hline
  $Q_{\max}$ & maximal strength of impulse & 0.06 Nms
  \\
  $P_1$ & strength of impulse of the 1st IC & $=Q_{\max}$
  \\
  $P_2$ & strength of impulse of the 2nd IC & $=-Q_{\max}$
  \\
  $\Delta \tau$ & width of impulse & $=\Delta t$
  \\
  $\tau_G$ & relaxation time of impulse generator &  $=\Delta t$
  \\ 
  $D$ & \multicolumn{2}{@{}l@{}}{%
      region of measurement \hfill
  \parbox[t]{12em}{%
  $[-0.13,0.43]\times[-3.28,10.58]$\\
  $\times[-0.35,0.31]\times[-3.80,5.15]$
  }
      }
  \\
  $\Delta t$ & step size of numerical integration &  $5\times10^{-4}$ s
  \\\hline
  \multicolumn{3}{l}{~}
  \\\hline
  \multicolumn{3}{l}{Free parameters} \\\hline
  $m$ & \multicolumn{2}{@{}l}%
        {resolution of numerical classifier \quad($m_j:=m$ for all $j$) }
	  \\
  $Q$ & \multicolumn{2}{@{}l}%
        {strength of impulsive disturbance}
      \\
  $\tau^d$ & \multicolumn{2}{@{}l}%
        {delay time of delay element}
	  \\
  \hline
 \end{tabular}
 \ltab{ICpara}
 \end{minipage}
\end{table}

\subsection{Individual performance of IC}

In order to investigate individual performance of IC, we first consider
impulse responses of the CIP model in \reqn{CIP8D} equipped with the IC in the
left-hand only given by
\begin{equation}
 \gv T = \Upd + \Uic + \gv v
  =\Upd +
  \begin{bmatrix}
   \uic_1\big(\gv x(t);\JJ_1\big) \\ 0
  \end{bmatrix}
  +
  \begin{bmatrix}
   v(t) \\ 0
  \end{bmatrix}
  ,\quad
  v(t):= Q\uimp(t)
  ,\quad
  \JJ_1:=\{2,3\},
  \leqn{IR}
\end{equation}
where $\Upd$ is the standing controller in \reqn{upd}, $Q$ is an impulse
of initial disturbance, $\uic_1$ is the four-dimensional IC developed in
\reqn{measure4D}, and $\uimp(t)$ is the unit impulse function defined
in \reqn{imp}. 

In the following numerical examples, we assume the maximal strength of
impulse $Q_{\max}=0.06$ so that $v(t)$ cannot produce switching motions
from the trivial initial state $\gv x(0)=\oo^{(8)}+w_0\ee^{(8)}_5$ (or
$\ep_1$) to the other stable states in order to avoid trivial switching
motions.
Following this assumption, the region of measurement $D$ is taken as
listed in \rtab{ICpara} so that it circumscribes at least all the
trajectories $\gv y(t)=H\gv x(t)$ for the maximal disturbance
$Q=Q_{\max}$. We also set the strength of impulse of IC to a common
value $P_1 = -P_2 = Q_{\max}$ to avoid the trivial switching motions
mentioned above.
Note that under the conditions of $k_w$ and $c_w$ listed in
\rtab{CIPpara} and for $Q\leq Q_{\max}$, dynamic change of length of the
connection rod in \reqn{IR} is limited to
\begin{math}
 |w(t)-w_0|/w_0 < 0.01
\end{math}
so that the measurement in \rsec{rigid} is expected to work.

Solving \reqn{CIP8D} with \reqn{IR} numerically from a trivial initial
state $\gv x(0)=\oo^{(8)}+w_0\ee^{(8)}_5$ for a given $Q$, we have the
corresponding final position $\ep_\nu$ and obtain correspondence from
$Q$ to $\nu$ as plotted in \rfig{IC50} for the resolution $m=50$ where
$Q$ is taken at $N_Q=100$ points with uniform increment over the interval
$[0,Q_{\max}]$.
The small circles represent the results of $\nu(Q)$ in presence of the
IC's outputs and the cross marks represent those in absence of the
outputs due to weak $v(t)$.

To evaluate the performance, we define a success rate in the following
form:
\begin{equation}
 E=E(\JJ):= N_\JJ/(N_Q-N_0)\quad (0\leq E\leq 1),
  \leqn{E}
\end{equation}
where $N_\JJ$ is the number of points on $Q$ at which
$\lim_{t\to\infty}\gv x(t)=\ep_\nu$ for all $\nu\in\JJ$ and $N_0$ is the
number of points in absence of the IC's outputs.  
The success rate of the result for $m = 50$ shown in \rfig{IC50} is calculated
as $E = 0.165$. Also, the rate for $m = 100$ can be obtained as $E =
0.305$ in the same manner. Therefore, it appears that our IC has low
performance.

\begin{figure}[t]
 \centering
 \includegraphics[width=0.7\hsize]{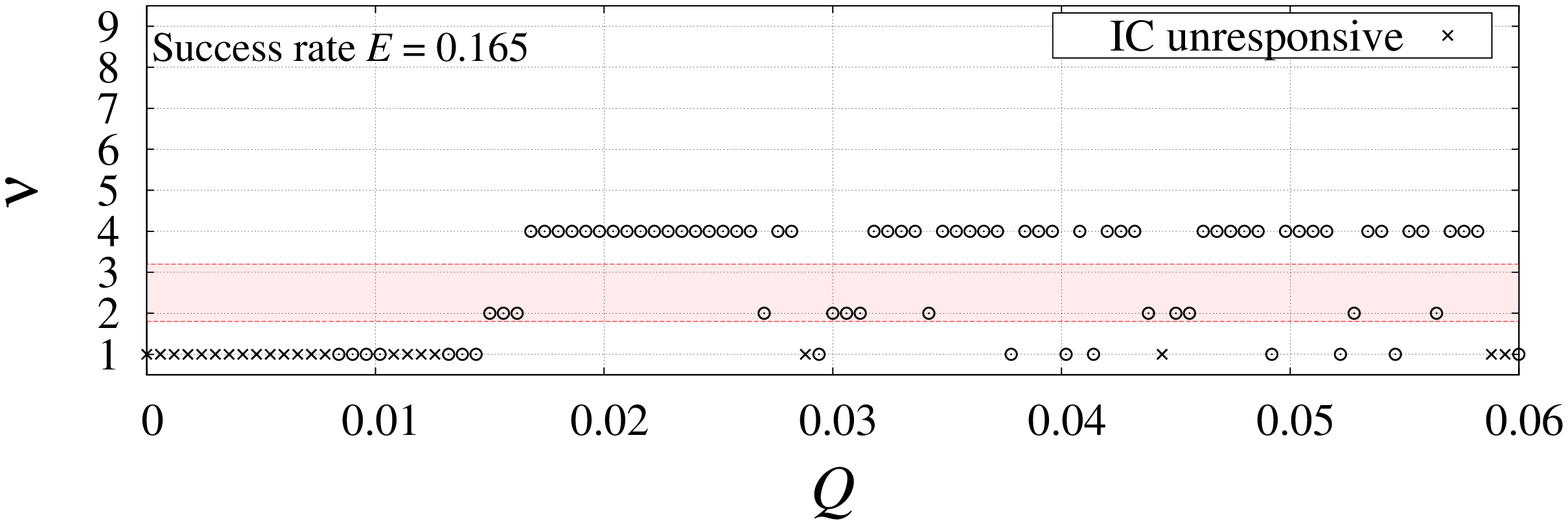}
 \caption{Index of final state $\ep_\nu$ as a function of the strength
 of initial disturbance $Q$ for the resolution $m = 50$.}
 \lfig{IC50}
 \par\vskip\floatsep
 \includegraphics[width=.4\hsize,trim=0 0 50 0]{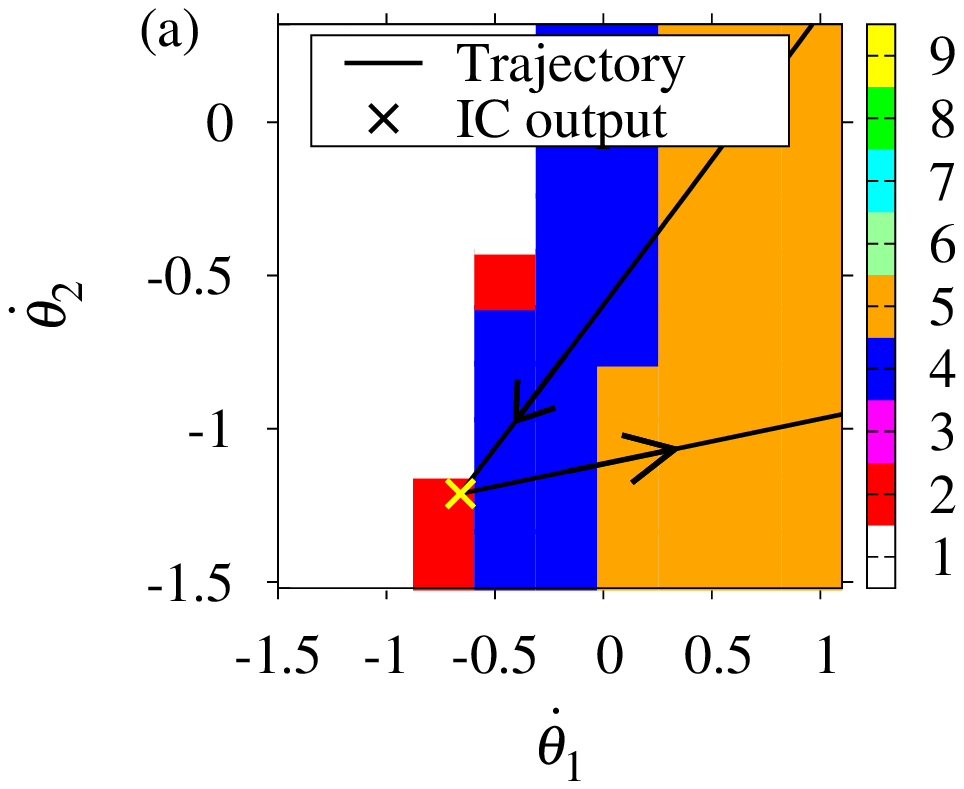}
 \hskip2em
 \includegraphics[width=.4\hsize,trim=0 0 50 0]{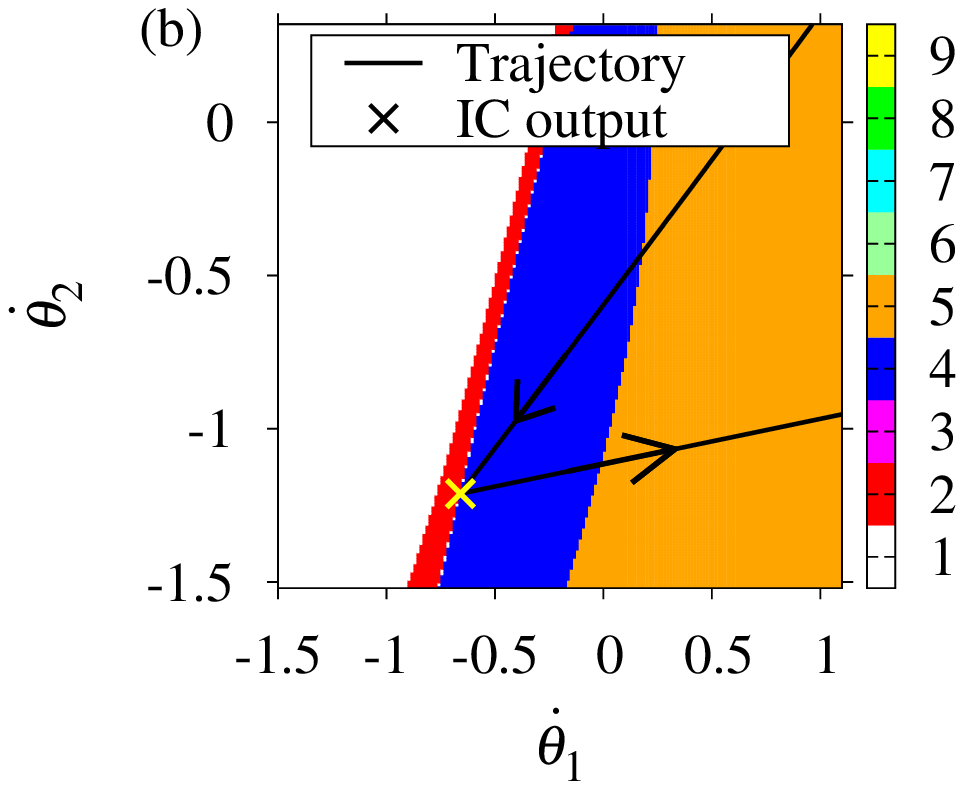}
 \caption{Misclassification of reachable sets. Colored areas are quantized reachable sets for (a) m=50 and (b) m=1000.}
 \lfig{howmiss}
 \par\vskip\floatsep
 \includegraphics[width=.6\hsize]{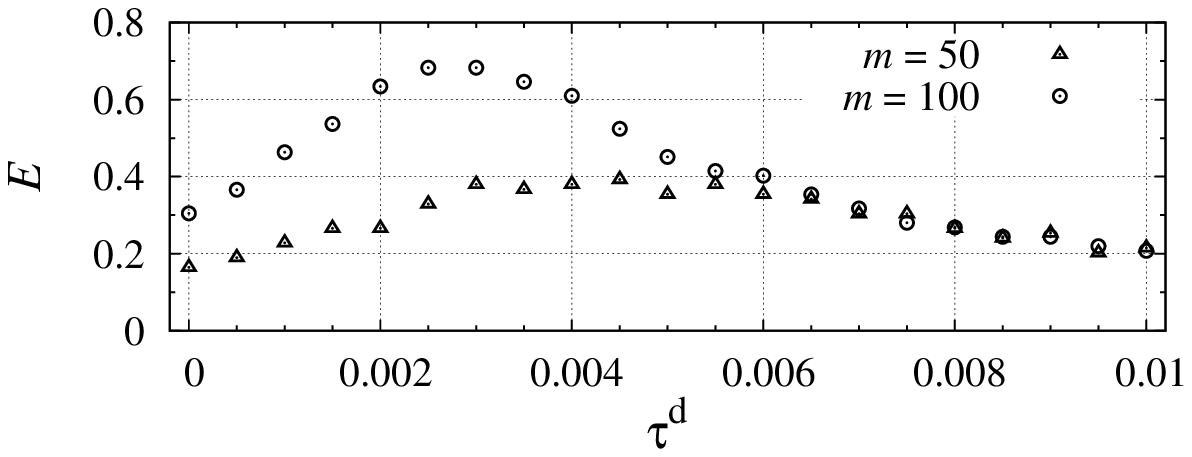}
 \caption{Success rate $E$ as functions of the delay time $\tau^d$ of
 DIC in \reqn{dic}.}
 \lfig{E:dic1}
\end{figure}

\subsection{Performance improvement with a delay element}

One of the reasons of the low performance above can be explained in
\rfig{howmiss}. The colored areas represent the quantized reachable sets:
\begin{equation}
 \Psi^{\ast}_\nu := \bigoplus
  \Big\{D^{\ii}\;\Big|\; \bar C(\ii) = \nu,\;\ii\in\II\Big\}
  \quad(\nu=1,\cdots,9),
  \leqn{Psi*}
\end{equation}
plotted on the hyper plane that contains the point $\gv x(t_1)$ on the
phase trajectory at which the IC's output occurs (indicated by the cross
mark). The resolutions of quantization of reachable sets are $m = 50$ for
\rfig{howmiss} (a) and $m = 1000$ for \rfig{howmiss} (b).
In the low resolution case in \rfig{howmiss} (a), it can numerically
be clarified that $\gv x(t_1)$ belongs to the reachable set
$\Psi^{\ast}_2$ (in red) of $\ep_2$ while $\gv x(t)$ ($t > t_1$) actually
converges to $\ep_4$ ($\neq\ep_2$). Such misclassification can be
refined by the high resolution as shown in \rfig{howmiss} (b). In this
case, the point $\gv x(t_1)$ is primly classified into $\Psi^{\ast}_4$
(in blue) of $\ep_4$. Although in theory, taking a sufficiently large
resolution $m$ provides nearly exact reachable sets, it greatly enlarges
computational efforts. Another approach to reduce the misclassification
is to replace the quantized reachable set $\Psi^{\ast}_i$ in \reqn{Psi*}
of $\ep_i$ with a subset $\Psi^{\circ}_i:= \Psi^{\ast}_i -
\Delta\Psi^{\ast}_i$ where $\Delta\Psi^{\ast}_i\subset\Psi^{\ast}_i$ is
a set of border points neighboring the other sets $\Psi^{\ast}_j$
($j\neq i$). However, extracting the border points is not necessarily
easy because the reachable sets sometimes exhibit nested structures as
discussed by one of the authors \citep{cip2009}. Actually, in \rfig{howmiss}
(b), quite narrow region of $\Psi^{\ast}_1$ (in white) appears between
$\Psi^{\ast}_2$ (in red) and $\Psi^{\ast}_4$ (in blue).

Therefore, let us take yet another approach by replacing the IC with a
delayed IC (DIC) in the following form:
\begin{equation}
 \udic_i\Big(\gv x(t);\JJ,\tau^d\Big)
  := \uic_i\Big(\gv x(t-\tau^d);\JJ\Big)
  \quad(i=1,2),
  \leqn{dic}
\end{equation}
where $\tau^d$ is a delay time. It seems reasonable to expect that a
trajectory about to crossing a course-grained border of a reachable set reaches
the true border soon. \Rfig{E:dic1} shows the success rate $E$
corresponding to this replacement given by
\begin{equation}
 \gv T 
  =\Upd +
  \begin{bmatrix}
   \udic_1\Big(\gv x(t);\JJ_1,\tau^d\Big) \\ 0
  \end{bmatrix}
  + \gv v
  ,\quad
  \JJ_1:=\{2,3\},
  \leqn{IR:dic}
\end{equation}
where $E$ is averaged over the two types of initial disturbance: $\gv
v(t)=\big(v(t),0\big)^T$ and $\big(0,v(t)\big)^T$, and the other
procedures of obtaining $E$ are the same as in \rfig{IC50}. In
\rfig{E:dic1}, the triangles and the circles represent $E$ as functions
of the delay time $\tau^d$ of DIC in \reqn{dic} for $m = 50$ and $100$
respectively. It is clearly seen that the functions $E(\tau^d)$ are
nearly concave down. The maximal values are $E = 0.392$ at
$\tau^d= 0.0045$ and $E=0.683$ at $\tau^d= 0.0025$ for $m = 50$ and $100$
respectively. Therefore, the DIC roughly doubles the success rate,
namely, $0.392/0.165\approx 2.38$ for $m = 50$ and $0.683/0.305\approx
2.24$ for $m = 100$.

\subsection{Competitive behavior}

In this final section, we install the DICs into both sides of
controllers as
\begin{equation}
 \gv T 
  =\Upd +
  \begin{bmatrix}
   \udic_1\Big(\gv x(t);\JJ_1,\tau^d_1\Big) \\
   \udic_2\Big(\gv x(t);\JJ_2,\tau^d_2\Big) 
  \end{bmatrix}
  + \gv v
  ,\quad
  \JJ_1:=\{2,3\}
  ,\quad
  \JJ_2:=\{4,7\},
  \leqn{comp:dic}
\end{equation}
where $\Upd$ is the standing controller in \reqn{upd}, $\udic_1$ and
$\udic_2$ is the four-dimensional DIC in \reqn{dic} through \reqn{uic2},
and $\gv v(t)=\big(v(t),0\big)^T$, $\big(0,v(t)\big)^T$ are initial
impulsive disturbances of strength $Q$ as shown in \reqn{IR}.
The competitive meanings of $\JJ_1$ and $\JJ_2$ are shown in \rfig{ep}.

\begin{figure}[t]
 \centering
 \begin{minipage}[b]{.53\hsize}
  \centering
  \includegraphics[width=\hsize]%
  {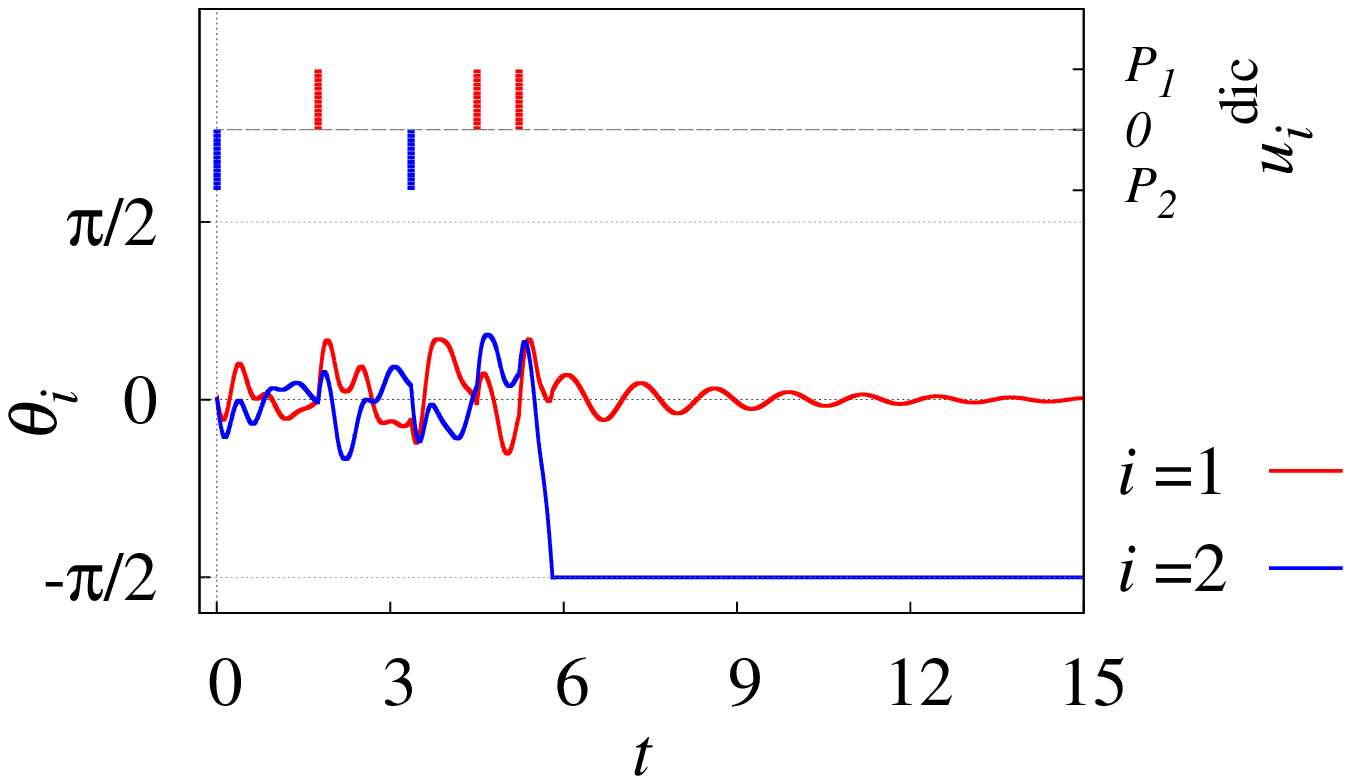}
  \par{(a) Time responses}
 \end{minipage}
 \hfill
 \begin{minipage}[b]{.45\hsize}
  \centering
  \includegraphics[width=\hsize]%
  {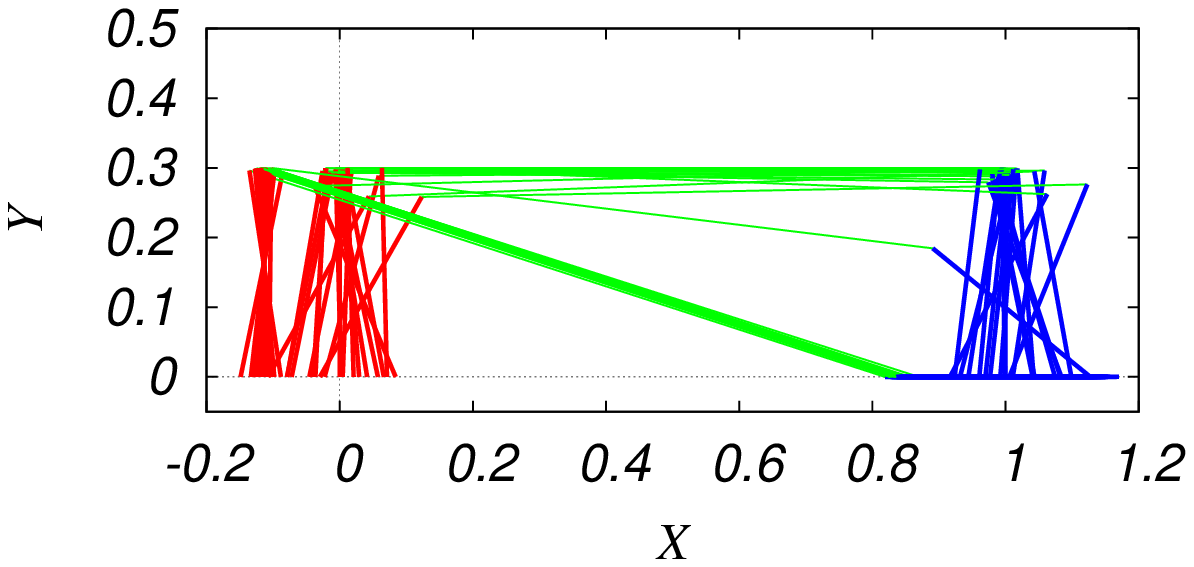}
  \par{(b) Motion of mechanism}
 \end{minipage}
 \caption{Competitive behavior for IC100 vs DIC50 ($\tau^d_2=0.0045$).}
 \lfig{mot:IC100vsDIC50}
 \par\vskip\floatsep
 \includegraphics[width=.7\hsize]{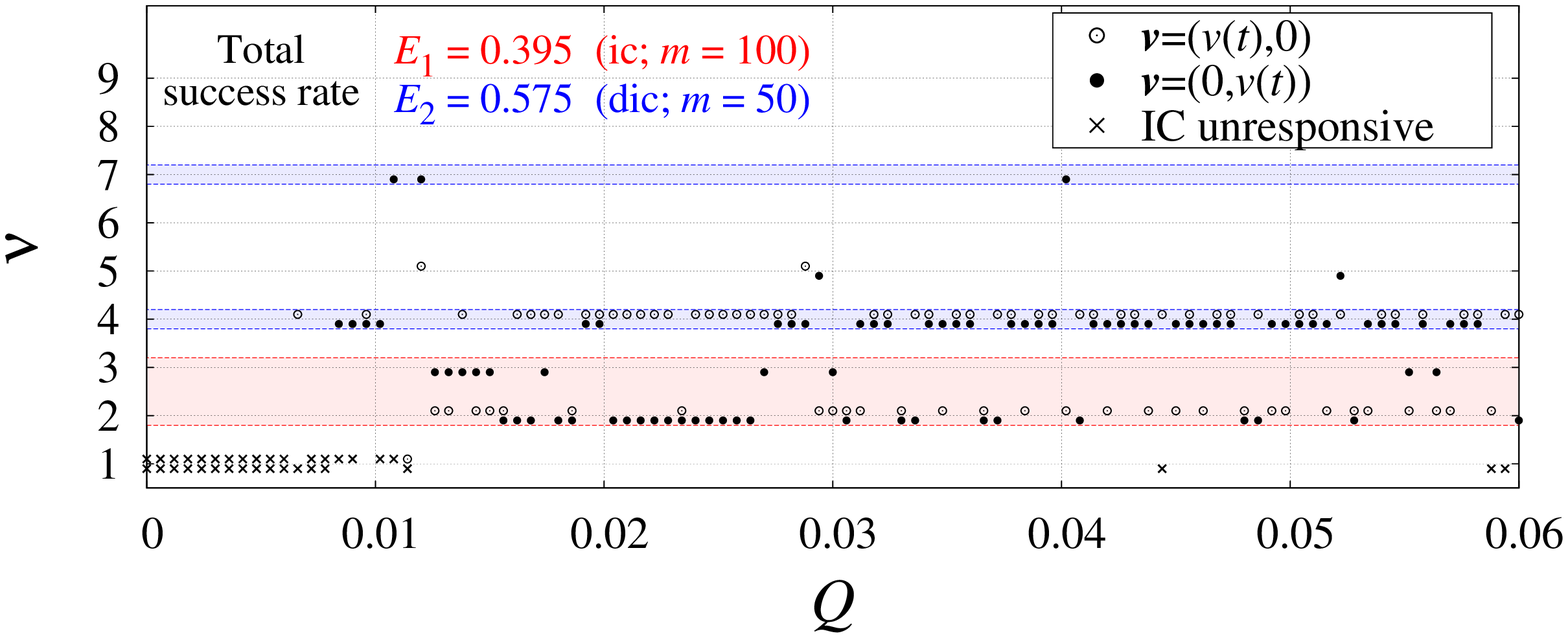}
 \caption{Index of final state $\ep_\nu$ as functions of 
 $Q$ for IC100 vs DIC50 ($\tau^d_2=0.0045$).}
 \lfig{IC100vsDIC50}
\end{figure}

\Rfig{mot:IC100vsDIC50} shows a competitive behavior between the normal IC:
\begin{math}
 \udic_1\big(\gv x(t);\JJ_1,0\big)
 =
 \uic_1\big(\gv x(t);\JJ_1\big)
\end{math}
for $m=100$ and the DIC with optimal delay time:
\begin{math}
 \udic_2\big(\gv x(t);\JJ_2,0.0045\big)
\end{math}
for $m=50$.
The time responses in \rfig{mot:IC100vsDIC50} (a) are obtained by solving
\reqn{CIP8D} with \reqn{comp:dic} from $\gv
x(0)=\oo^{(8)}+w_0\ee^{(8)}_5$ under the conditions listed in
\rtab{CIPpara} and \rtab{ICpara} by applying the initial disturbance
$\gv v(t)=\big(v(t),0\big)^T$ with $Q=0.0186$.
The same solution is plotted in \rfig{mot:IC100vsDIC50} (b) as a motion
of the CIP mechanism. For convenience, we refer to the former IC as 
``IC100'' and the later as ``DIC50''.

It is shown in \rfig{mot:IC100vsDIC50} that the ICs above generate
the impulsive control forces autonomously to make the CIP system drop
into $\ep_2$.  According to the competitive interpretation in \rfig{ep}
inspired by some kind of wrestling match, it can be said that ``the left
agent wins by pulling the right agent.''

\Rfig{IC100vsDIC50} shows the result of competition. The plots are
obtained in the same manner as those in \rfig{IC50}, except that in
\rfig{IC100vsDIC50}, the open and filled circles represent
the results for 
$\gv v(t)=\big(v(t),0\big)^T$ and $\big(0,v(t)\big)^T$
respectively, and that
the cross marks represent the results in which neither of IC100 and
DIC50 produces its own output. Similar to the individual
case in \reqn{E}, we define the success rate of this competition as
\begin{equation}
 E_i=E(\JJ_i):= N_{\JJ_i}/(N_Q-N_0)
  \quad (0\leq E_i\leq 1)
  \quad (i=1,2),
\end{equation}
where $N_Q = 100 \times 2$ is the number of all plots in
\rfig{IC100vsDIC50}, $N_0$ is the number of trials in absence of the
IC's outputs (cross marks), and the definition of $N_{\JJ_i}$ is the
same as that of $N_{\JJ}$ in \reqn{E}.  From the result in
\rfig{IC100vsDIC50}, the success rates are obtained as $E_1 = 66/167
\approx 0.395$ for IC100 and $E_2 = 96/167 \approx 0.575$ for DIC50. 
Therefore, it is shown that at least based on the present definition of
competition and success rate, the performance improvement using
time delay is more effective than that doubling the quantization
resolution without the time delay.

\section{Conclusion}

In this paper, we discussed a competitive problem in which mechanical
agents are fighting with each other and formulated it as the set of: (A)
the nonlinear dynamical model with nine stable equilibriums and (B) the
matrix describing competitive interpretation of these
equilibriums. Based on this framework, we proposed a competitive IC that
receives the state vector and output the impulsive forces to make the
competitor fall down. Developing a quantized and reduced order design of
the controller, we derived a practical control procedure along with an
off-line learning method. To investigate performance of the
controller in individual use and also in competition use, we conducted
numerical experiments and obtained the following results:

\begin{itemize}
 \item The individual performance of IC depends on resolutions of the
       quantized reachable sets.
 \item The individual performance of IC can be improved by adding the
       delay element into the IC.
 \item To improve the competitive performance of IC, adding the delay
       element may become more effective than refining the resolutions.
\end{itemize}

In future work, we plan to investigate a further order reduction of
measurement based on time delayed embedding methods and to improve
classification accuracy by applying machine learning techniques. We also
plan to conduct competitions between humans and the proposed ICs.
Moreover, application of position control to the carts will
be considered to investigate competitive problems in a bounded area.

\section*{Acknowledgment}

The authors appreciate the feedback offered by Dr. Munehisa Sekikawa.


\end{document}